\documentclass{ecai}
\usepackage{times}
\usepackage{graphicx}
\usepackage{latexsym}

\usepackage{algorithm}
\usepackage{algpseudocode}
\usepackage{pifont}
\usepackage{amsmath}
\usepackage{bbm}
\usepackage{float}

\begin{document}
\newcommand\floor[1]{\lfloor#1\rfloor}
\newcommand\ceil[1]{\lceil#1\rceil}

\title{Online Adaptation of Deep Architectures with Reinforcement Learning}

\author{Thushan Ganegedara, Lionel Ott \and Fabio Ramos \institute{University of Sydney, Australia email: tgan4199@uni.sydney.edu.au, lott4241@uni.sydney.edu.au and fabio.ramos@sydney.edu.au} }

\maketitle
\bibliographystyle{ecai}

\begin{abstract}
Online learning has become crucial to many problems in machine learning. As more data is collected sequentially, quickly adapting to changes in the data distribution can offer several competitive advantages such as avoiding loss of prior knowledge and more efficient learning. However, adaptation to changes in the data distribution (also known as covariate shift) needs to be performed without compromising past knowledge already built in into the model to cope with voluminous and dynamic data. In this paper, we propose an online stacked Denoising Autoencoder whose structure is adapted through reinforcement learning. Our algorithm forces the network to exploit and explore favourable architectures employing an estimated utility function that maximises the accuracy of an unseen validation sequence. Different actions, such as \emph{Pool}, \emph{Increment} and \emph{Merge} are available to modify the structure of the network. As we observe through a series of experiments, our approach is more responsive, robust, and principled than its counterparts for non-stationary as well as stationary data distributions. Experimental results indicate that our algorithm performs better at preserving gained prior knowledge and responding to changes in the data distribution.
\end{abstract}

\section{Introduction}
%

Over the past decade, Deep Architectures ~\cite{hinton2006fast},~\cite{bengio2007greedy} have become a widely-discussed topic in machine learning. One key reason being the ability to jointly perform feature-extraction and classification on raw data, outperforming many other techniques in various domains including object recognition ~\cite{krizhevsky2012imagenet},~\cite{cirecsan2012multi}, hand-writing recognition ~\cite{hinton2006fast} and speech recognition ~\cite{hinton2012deep}. A deep network can be understood as a neural network consisting of many hidden layers ~\cite{deng2014deep}. While the interest in deep networks arose quite early, only the recent hardware and optimisation developments (e.g. Graphical Processing Units (GPUs), Greedy pre-training) sparked the practicality of deep architectures.
 
Despite the note-worthy learning capacity, deep architectures are still susceptible to the past-knowledge being overridden due to \emph{Covariate Shift}~\cite{sugiyama2012machine}. Covariate shift is a common phenomenon that transpires in online settings. Covariate shift essentially refers to the difference in training and testing data distributions. Successful exploitation of adaptive capabilities of deep networks to minimise the adverse effects of the covariate shift will lead to new frontiers in data science. 

While many algorithms (especially Support Vector Machines (SVM)) have been enhanced with online learning capabilities ~\cite{liu2008incremental},~\cite{poggio2001incremental}, only few attempts of incorporating online learning for Neural Networks have been proposed in the literature, notably in ~\cite{zhou2012online},~\cite{riedmiller2005neural},~\cite{mnih2013playing}, and~\cite{stanley2002evolving}. Of these, only~\cite{stanley2002evolving} and~\cite{zhou2012online} focus on changing the structure of the network, where the others focus on adapting a fixed architecture accordingly.~\cite{stanley2002evolving} proposes an intriguing approach to evolve neural networks using genetic algorithm, by mutating weights and nodes in the network and crossing over existing networks to generate more fit off-springs. However, this technique is not scalable for deep networks and requires many repetitive runs through the data.~\cite{zhou2012online} proposes a structural adaptation technique for deep architectures relying on simple heuristic (i.e. immediate performance convergence). ~\cite{zhou2012online} does not seek a long-term reward and lacks in responsiveness, as it waits for a pool of data to be filled in order to add nodes to the structure. These limitations motivate the question of how to explore the space of different architectures in an online setting in a more responsive, robust and principled manner. 

In this paper, we introduce a state-of-the-art mechanism to modify deep architectures (specifically Denoising Autoencoders ~\cite{vincent2010stacked}) based on reinforcement learning. The decision making behaviour exploits and explores possible actions to discover favourable modifications to the structure (i.e. adding/removing nodes) by maximising a stipulated reward over time. Adding nodes helps to accommodate new features, while removing nodes helps to remove redundant features. An additional pooling operation fine-tunes the network with previously observed data. The method keeps track of a continuously updated utility (long-term reward) function to decide which action is best for a given state, whose estimation will improve over time. The experimental results on three datasets clearly show that our algorithm outperforms its counterparts in both stationary and non-stationary situations.
\vspace{-0.3cm}
\section{Background}
\subsection{Online Learning}

By online learning we refer to the ability to accommodate new knowledge (i.e. features) without overriding previously acquired knowledge (i.e. features) ~\cite{thrun2012learning}. This is becoming more popular due to the explosive growth of data. Online learning has the ability to learn from a continuous stream of data without a loss of past knowledge and attempts to address the non-stationary nature of data by allowing more flexibility in the model. For this reason, online algorithms perform significantly better in handling problems with covariate shift.

\subsection{Deep Networks}
We begin the presentation of the method by introducing the following notation:  
\begin{itemize}
\item $\mathbf{x}$ - Inputs
\item $\mathbf{y}$ - Input labels
\item $K$ - Number of classes
\item $\mathbf{\tilde{x}}$ - Noise-corrupted input
\item $\mathbf{\hat{x}}$ - Reconstructed input
\item $W$ - Weights of a neuron layer
\item $b$ - Bias of a neuron layer
\item $b'$ - Reconstruction bias of a neuron layer
\end{itemize}

\subsubsection{Autoencoder}
An Autoencoder ~\cite{hinton2006reducing}  maps a set of inputs $\mathbf{x}=$\{$\mathbf{x}_i \in [0,1]^D$\} $\forall i=1,...,N$ where $\mathbf{x_i}=\{{x_i^1,x_i^2,...,x_i^D}\}$ and $D$ is dimensionality of data to a latent feature space $H$ with $h_{W,b}(\mathbf{x}) = sig(W\mathbf{x}+b)$, where $W \in {\rm I\!R}^{H\times D}$, $b \in {\rm I\!R}^H$ and $sig(s) = \frac{1}{1+\exp{-s}}$. An autoencoder can reconstruct the input $\mathbf{\hat{x}}_i$ $\forall i=1,...,N$ from the latent feature space $H$ with $\mathbf{\hat{x}} = sig(W^T \times h_{W,b}(\mathbf{x})+b')$ where superscript $^T$ denotes transpose and $b' \in {\rm I\!R}^D$. For simplicity we assume tied weights.

\subsubsection{Denoising Autoencoder}
The Denoising Autoencoder (DAE) is a variant of autoencoder which uses a corrupted (noisy) version of the example as the input ~\cite{vincent2010stacked}. This forces the algorithm to become more robust to noise. DAE works in the following manner.

First, the inputs are corrupted by introducing noise using a binomial distribution with probability $p$. Let us call the corrupted input $\mathbf{\tilde{x}}$. Next, $\mathbf{\tilde{x}}$ is mapped to a hidden representation using $h_{W,b}(\mathbf{\tilde{x}}) = sig(W\mathbf{\tilde{x}}+b)$ where $W \in {\rm I\!R}^{H\times D}$, $b \in {\rm I\!R}^H$ and $sig(s) = \frac{1}{1+\exp{-s}}$. Finally, the decoding function retrieves the reconstructed input, $\mathbf{\hat{x}} = sig(W^T \times h_{W,b}(\mathbf{\tilde{x}})+b')$, where $b' \in {\rm I\!R}^D$. In this work, we assume tied weights for encoding and decoding. \emph{Cross entropy} is used as the cost function (Equation ~\ref{eq:l_gen}),
\begin{equation} \label{eq:l_gen}
L_{gen}(\mathbf{x},\mathbf{\hat{x}}) = \sum_{j=1}^{D}x^j\text{log}(\hat{x}^j) + (1-x^j)\text{log}(1-\hat{x}^j).
\end{equation}
The optimal values for parameters $W,b,b'$ are found by minimising the cost function, 
$$W_{opt},b_{opt},b'_{opt} = \text{argmin}_{W,b,b'}L_{gen}(\mathbf{x},\mathbf{\hat{x}}).$$

\subsubsection{Stacked Denoising Autoencoders}
A Stacked Denoising Autoencoder (SDAE) ~\cite{vincent2010stacked} is a set of connected autoencoders. A SDAE undergoes two main processes; pre-training and fine-tuning. In the pre-training process, the network is considered as a set of autoencoders $AE^1,...,AE^L$. The output of $AE^l=\{W^l,b^l,b'^l\}$, $h_{W,b}^l$ where $l$ is the current layer, is calculated as follows,

\[
  h_{W,b}^l=\begin{cases}
               sig(W^l\mathbf{\tilde{x}}+b^l); \hspace{.5em} \text{if} \hspace{.5em} l= 1\\
               sig(W^lh_{W,b}^{l-1}+b^l); \hspace{.5em} \text{Otherwise.}                
            \end{cases}
\]

In the fine-tuning phase, the network is treated as a single deep autoencoder and trained using labelled data $\mathcal{D}$. Assuming labelled data in the format $\mathcal{D} = (\mathbf{x_i},\mathbf{y_i} )$, $ \forall i=1,...,N$ where $\mathbf{y_i} \in \{0,1\}^K$ such that if $y_i^j$ are the elements of $\mathbf{y}_i$ then $\sum_j{y_i^j}=1$, we can use a \emph{softmax} layer with parameters \{$W^{out}$, $b^{out}$,$b'^{out}$\}. The output of the network is defined as $\mathbf{\hat{y}} = \text{softmax}(W^{out}h_{W,b}^L+b^{out})$, where softmax($a_k$) = $\frac{\exp(a_k)}{\sum_{k'}\exp(a_{k'})}$. Then the cost function becomes,
\begin{equation} \label{eq:l_disc}
L_{disc}(\mathbf{y},\mathbf{\hat{y}}) = \sum_{j=1}^K(y^j\text{log}\hat{y}^j + (1-y^j)\text{log}(1-\hat{y}^j)).
\end{equation}
Finally, from Equation ~\ref{eq:l_disc}, we can formulate the optimisation problem to learn $W$, $b$
and $b'$ as,
$$W_{opt},b_{opt},b'_{opt} = \text{argmin}_{W,b,b'}L_{disc}(\mathbf{y},\mathbf{\hat{y}}), $$ 
where $W_{opt}$=$(W^1_{opt},...,W^L_{opt},W^{out}_{opt})$, $b_{opt}$=$(b^1_{opt},...,b^L_{opt},$ $b^{out}_{opt})$ and $b'_{opt}$=$(b'^1_{opt},...,b'^L_{opt},b'^{out}_{opt}).$

\subsection{Incremental Feature Learning for Denoising Autoencoders}
\label{sub_sec_mincdae}

Merge-Incremental Denoising Autoencoders (MI-DAE) is an online learning stacked denoising autoencoder proposed in~\cite{zhou2012online}. Initially, the network is pre-trained using a pool of data (typically first 12,000 examples). Then, for every batch of data $b_t$, add hard examples (i.e. $\mathbf{x_i}$ if $ L_{gen}(\mathbf{x_i}, \mathbf{\hat{x}_i}) > \frac{\sum_{\forall \mathbf{x_j} \in b_t}L_{gen}(\mathbf{x_j},\mathbf{\hat{x}_j})}{|b_t|}$) to a pool, $B$. The method then performs merging of nodes within the same layer or adds new nodes to the network.  
Once the number of points in $B$ exceeds a threshold, $\tau$, retrieve previously calculated pairs of nodes with the highest similarity ($\Delta$Mrg) and add $\Delta$Inc new nodes to the network. Next, use $B$ to greedily train newly added features. Afterwards, update $\Delta$Mrg and $\Delta$Inc ~\cite{zhou2012supplementary} and remove all data from $B$. Finally repeat this process for all the batches in the sequence. Pseudo-code for this algorithm
is presented in Algorithm ~\ref{algo_mergeinc}.

\begin{algorithm}
\caption{MergeInc Algorithm}
\label{algo_mergeinc}
\begin{algorithmic}[1]
\Procedure{MergeInc($b_t$,$\Delta$Mrg,$\Delta$Inc)}{}
\State \textbf{Define}: $\mu$ - Average reconstruction error for the 
\State \hspace{1.2cm} most recent 10,000 examples 
\State \textbf{Define}: $\tau$ - Pool threshold (10,000 examples) 
\State  Compute objective $L_{disc}(\mathbf{y_j},\mathbf{\hat{y}_j})$, $\forall$ $\{\mathbf{x_j},\mathbf{y_j}\}$ $\in$ $b_t$
\State Add hard example $\mathbf{x_j}$ to $B$ if $L_{gen}(\mathbf{x_j},\mathbf{\hat{x}_j}) > \mu$ of $b_t$
\hspace{\algorithmicindent} \hspace{\algorithmicindent} \If{$|B| > \tau $}
    \State  Merge $2\Delta$Mrg candidates to $\Delta$Mrg
    \State Add $\Delta$Inc nodes and fine-tune $\Delta$Inc new nodes with
    
    \hspace{0.15cm} $\{\mathbf{x_j},\mathbf{y_j}\}$ $\in$ $B$ while keeping rest of the network constant
    \State Update $\Delta$Mrg and $\Delta$Inc (Heuristic-based ~\cite{zhou2012supplementary})
    \State Set $B=\emptyset$
\hspace{\algorithmicindent} \hspace{\algorithmicindent} \EndIf
\State Fine-tune all the features (with $\Delta$Mrg and $\Delta$Inc) with $b_t$
\EndProcedure
\end{algorithmic}
\end{algorithm}

 \begin{table*}[t]
\centering
\caption{The notations and definitions used in Section ~\ref{sec_ra_dae}}
\begin{tabular}{ |c|p{7cm}||c|p{7cm}|  }
 \hline
Notation & Description & Notation & Description\\
  \hline
$N$ & Number of data points &$B_r$ & Pool of data containing most recent $\tau$ examples\\
$D$ & Dimensionality of data & $B_{ft}$ & Pool of data containing dissimilar inputs\\
$K$ & Number of classes & $\Lambda$ & Distance threshold for $B_{ft}$\\
$p$ & Number of data points in one batch & $\mathcal{\tilde{L}}^n(m)$ & Exponential Moving Average of error L in the window $n-m$ to $n$\\
$n$ & sequence number of the current batch of data & $\nu_l^n$ & Ratio between the current count of neurons and the initial count for neuron layer $l$ for $n^{th}$ data batch\\
$\tau$ & Size of data pools & $\Delta \text{Inc}$ & Number of neurons to add at a given time\\
$\mathbf{x}_i$ & $i^{th}$ data point & $\Delta \text{Mrg}$ & Number of neurons to remove at a given time\\
$\mathbf{y}_i$ & Vectorized label of $\mathbf{x}_i$ s.t $\forall$ $y_i^j\in \mathbf{y}_i$ $y_i^j \in \{0,1\}$ s.t. $\sum_j{y_i^j}=1$ & $r^n$ & The reward for the $n^{th}$ batch of data\\
$\mathcal{D}$ & Dataset containing $\{\{\mathbf{x}_1,\mathbf{y}_1\},\{\mathbf{x}_2,\mathbf{y}_2\},\ldots\}$ &$\gamma$ & Discount rate for Q value update\\
$\mathcal{D}^n$ & $n^{th}$ batch of data ($\mathcal{D}^n \subset \mathcal{D}$) &$Q(s,a)$ & Utility function\\
$L_g^n$ & Generative error for $n^{th}$ batch of data & $\eta_1$ & The duration until beginning to collect state-action pairs\\
$L_c^n$ & Classification error for $n^{th}$ batch of data & $\eta_2$ & The duration until beginning to exploit Q-values\\
\hline
\end{tabular}
\label{tbl_def}
\end{table*}

\subsection{Reinforcement Learning and Markov Decision Processes}
After describing SDAE, we now introduce notation and the basics of reinforcement learning (RL). RL enables an agent to learn a \emph{policy}, $\pi$ (a function that defines which action to take in a given state), by interacting with its environment, preferably trading-off between exploration and exploitation. A reinforcement learning task that satisfies the \emph{Markov Property} can be formulated as a Markov decision process (MDP)~\cite{sutton1998reinforcement}. Formally a Markov Decision Process can be defined using the following,
\begin{itemize}
	\item A set of states - S
	\item A set of actions - A
	\item A transition function - $T : S \times A \times S \to [0,1]$
	\item A reward function - $R : S \times A \times S \to {\rm I\!R}$.
\end{itemize}
In this paper, RL is used to find the policy to adapt the structure of the network, given the current network configuration or state. Therefore, at a given instance $i$, from state $s^i$ an action $a^i$ is performed and the network transits to state $s^{i+1}$. Actions are modifications to the network such as adding new nodes or removing existing nodes. The state is a function of the network performance and will be defined in Section~\ref{sec_ra_dae}. The reward $r^i$ for going from state $s^i$ to $s^{i+1}$ by taking action $a^i$ is calculated based on the errors produced on the learning task. State $s^{i+1}$ depends on the current state $s^i$ and current action $a^i$, and is conditionally independent of all the previous states and actions, thus satisfying the \emph{Markov Property}. The ultimate goal is to learn an optimal policy $\pi^*(s^i,a^i)$ that recommends the best action $a^i$ for a given state $s^i$.

In order to learn the policy to select the best action for a given state, Q-Learning is used. Q-Learning (a variant of Temporal difference ~\cite{sutton1998reinforcement}) is an off-policy model-free approach to finding the optimal policy, $\pi^*$. Q-Learning estimates the utility value in an online manner and, as an off-policy learning, it learns a value function independent of the agent's experience. This leads to exploring new tactics the agent has not tried. Furthermore, Q-Learning can be employed for MDPs with unknown transition and reward functions. Q-Learning proceeds as follows, 
\begin{enumerate}
\item Define $Q(s^n,a^n)$, where $s^n \in S$ and $a^n \in A$.
\item Initialise $Q^0(s^i,a^i) = 0$, $\forall s^i \in S$ and $\forall a^i \in A$.
\item Update $Q^{t+1}(s^n,a^n) = (1-\alpha)\times Q^{t}(s^n,a^n) + \alpha \times [R(s^n,a^n,s^{n+1}) + \gamma (max_{a'} (Q(s^{n+1},a')))]$ where $\gamma$ is the discount rate, $\alpha$ is the learning rate, and $s^{i+1}$ is the state after action $a^i$.
\end{enumerate}
One of the applications of using Q-learning is to train a multi-layer perceptron as found in ~\cite{riedmiller2005neural}. More recently, a variant of Q-Learning was successfully used in a Convolutional Deep Network when the network was trained to play the \emph{Atari} games using raw pixel images ~\cite{mnih2013playing}.

\vspace{-0.1cm}
\section{Reinforced Adaptive Denoising Autoencoder (RA-DAE)}
\label{sec_ra_dae}

\subsection{Limitations of MI-DAE}
MI-DAE (Algorithm \ref{algo_mergeinc}) introduces several interesting concepts useful for online learning such as, pooling data and update rules for $\Delta$Mrg and $\Delta$Inc. However, the approach has several limitations: (1) The response of the algorithm to changes is slow as it waits for a pool of data ($B$) to be filled in order to execute an operation; (2) While the algorithm incorporates an intuitive criteria (performance convergence) to modify the network (update rules), the method is based on simple heuristics such as the immediate future reward that does not generally reflect a holistic view of the effect an action has on the network. 

\vspace{-0.1cm}
\subsection{Overview of RA-DAE}
Motivated by the drawbacks in MI-DAE, we propose a more robust and principled solution which relies on RL. In essence, our algorithm estimates an utility function $Q(s,a)$ for each state-action pair by sampling from the environment, where actions are modifications in the network structure. Using $Q(s,a)$, the algorithm selects the best action for a given state. The utility function is based on the accuracy measured on an unseen validation batch. Our approach is beneficial as, 
\begin{itemize}
\item Actions are taken for every batch of data, resulting in fast response to sudden changes in the data distribution;
\item The utility function ensures that actions are taken based on the long-term benefit they incur on the accuracy;
\item A new pool operation refreshes the network's knowledge by fine-tuning the network using a pool of data containing data points \emph{significantly} different from each other.
\end{itemize}
 
\noindent\textbf{Notation}: 
An input data stream is denoted as $\mathcal{D} =\{\{\mathbf{x}_1,\mathbf{y}_1\},\{\mathbf{x}_2,\mathbf{y}_2\},\ldots\}$, where $\mathbf{x}_i$ is a normalized data point, $\mathbf{x}_i\in[0,1]^D$, $\mathbf{y}_i \in \{0,1\}^K$ and $y_i^j$ are the elements of $\mathbf{y}_i$ with $\sum_j{y_i^j}=1$. The $n^{th}$ data batch is written as $\mathcal{D}^n=\{\{\mathbf{x}_{(n-1) \times p}$ $,\mathbf{y}_{(n-1) \times p}\},...,\{\mathbf{x}_{n\times p},\mathbf{y}_{n\times p}\}\}$, where $p$ is the number of examples per batch. Denote the generative error as L$_{g}^n=\frac{\sum_{\forall \mathbf{x_i}\in \mathcal{D}^n} L_{gen}(\mathbf{x_i},\mathbf{\hat{x_i}})}{p}$  and the classification (or discriminative) error as L$_{c}^n=\frac{\sum_{\forall \mathbf{y_i} \in \mathcal{D}^n} \mathbbm{1}_{\hat{k_i}=k_i}}{p}$, where $\mathbbm{1}$ is the \emph{indicator} function and $k_i=argmax_{k'}(\{y_i^{k'}\})$,  $\forall k'=1,...,K$ of the $n^{th}$ batch. $r^n$ denotes the reward for the $n^{th}$ batch. Finally define two pools $B_{r}=\{\mathcal{D}^{n-\tau},...,\mathcal{D}^n\}$ and 
\begin{equation} \label{eq:b_ft}
\hspace{-0.06cm}
B_{ft}=\begin{cases}\mathcal{D}^n \hspace{0.2cm} \text{if} \hspace{0.2cm} B_{ft}=\emptyset \\
\mathcal{D}^n \cup B_{ft} \hspace{0.2cm} \text{if} \hspace{0.2cm}  d(\mathcal{D}^n,\mathcal{D}^j)>\Lambda \hspace{0.2cm} \forall \mathcal{D}^j \in B_{ft}\\
B_{ft}-\mathcal{D}^j \hspace{0.1cm} j=argmin_{j'}(\forall \mathcal{D}^{j'} \hspace{-0.15cm}\in \hspace{-0.05cm} B_{ft})\hspace{0.1cm} \text{if} \hspace{0.1cm} |B_{ft}|>\tau  \\
B_{ft} \hspace{0.2cm} \text{otherwise}
\end{cases} 
\end{equation}
for some $d$ distance measure and a similarity threshold $\Lambda \in [0,1]$. $\eta_1$ and $\eta_2$ are pre-defined thresholds for starting to collect observed state-action pairs and exploiting Q-values respectively. $\alpha$ is the learning rate for Q-learning. A summary of the notation is in Table ~\ref{tbl_def} for quick reference.

\subsection{RL Definitions}
To calculate when and which actions to take, we employ a MDP formulation. We define a set of states $S$, a set of actions $A$, and a reward function $r^n$ below. 

\subsubsection{State Space}
\label{sub_sub_sec_state_space}
The state space $S$ is defined as follows. For the $n^{th}$ batch,
\begin{align} \label{eq:cont-states}
S=\{\mathcal{\tilde{L}}_{g}^{n}(m), \mathcal{\tilde{L}}_{c}^n(m), \nu_1^n \} \in {\rm I\!R}^3
\end{align}
where the moving exponential average ($\mathcal{\tilde{L}}$) is defined as $\mathcal{\tilde{L}}^n(m) = \alpha \text{L}^n + (1-\alpha) \mathcal{\tilde{L}}^{n-1}(m-1)$, $n \geq m$ and $m$ is a pre-defined constant. $\mathcal{\tilde{L}}_{g}$ and $\mathcal{\tilde{L}}_{c}$ denote $\mathcal{\tilde{L}}$ w.r.t. L$_{g}$ and L$_{c}$, respectively, and  $\nu_l^n = \frac{\text{Node Count}_{current}}{\text{Node Count}_{initial}}$ for the $l^{th}$ hidden layer. $\mathcal{\tilde{L}}$ is defined in terms of recursive decay to respond rapidly to immediate changes.

This state space takes into account the following attributes:
\begin{itemize}
\item Ability of RA-DAE to classify an unseen batch of data;
\item Difference between current data distribution and previously observed distributions;
\item Complexity of RA-DAE's current structure.
\end{itemize}
The justification for the choice of state space is discussed in Section \ref{sub_sub_sec_eval_state}.

\subsubsection{Action Space}

\noindent The actions space is defined as,
\begin{equation} \label{eq:actions}
A = \{Pool, Increment(\Delta \text{Inc}), Merge(\Delta \text{Mrg})\},
\end{equation}
where $Increment$($\Delta$Inc) adds $\Delta$Inc new nodes and greedily initialise them using pool $B_{r}$. The $Merge$($\Delta$Mrg) operation is performed by merging the 2$\Delta$Mrg nodes. \emph{Merge} operation is executed by selecting the closest pairs (e.g. minimum Cosine distance) of $\Delta$Mrg nodes and merging each pair to a single node. The $Pool$ operation fine-tunes the network with $B_{ft}$. Both operations (i.e. Increment and Merge) are performed in the 1\textsuperscript{st}  hidden layer. Equations~\ref{eq:change}, ~\ref{eq:inc} and ~\ref{eq:mrg} outline the calculations for $\Delta$Inc and $\Delta$Mrg,
\begin{align} \label{eq:change}
\Delta = \lambda \exp^{\frac{-(\nu-\hat{\mu})}{2\sigma^2}}|L^n_{c} -L^{n-1}_{c}|
\end{align}
\begin{equation} \label{eq:inc}
  \Delta \text{Inc}=\begin{cases}
               \Delta; \hspace{.5em} \text{if} \hspace{.5em} a= Increment\\ 
               0; \hspace{.5em} \text{Otherwise}
            \end{cases}
\end{equation}
\begin{equation} \label{eq:mrg}
  \Delta \text{Mrg}=\begin{cases}
               \Delta; \hspace{.5em} \text{if} \hspace{.5em} a= Merge\\ 
               0; \hspace{.5em} \text{Otherwise}   
            \end{cases}
\end{equation}
where $\lambda$ is a coefficient controlling the amount of change, $\hat{\mu}$ and $\sigma$ are chosen depending on how large or small the network is allowed to grow, and $a$ is the current action chosen by Algorithm \ref{algo_qlearn}.  We defined $\Delta$Mrg and $\Delta$Inc as a function of $\nu_1^n$ and L$_{c}^n$. The objective of Equation \ref{eq:change} is to minimise the error while preventing the network from growing too large or too small. For example, if the error is high, the algorithm increases $\Delta$ to reduce the error. If the error has converged, i.e. has not changed for two consecutive batches, $\Delta$ will be small. 

The need for two pools, $B_{r}$ and $B_{ft}$ is justified as follows. The pool operation is designed to revise the existing knowledge. Thus, $B_{ft}$ is composed of a diverse set of data batches that differ in the distribution of the data. The objective of the increment operation is to add the most recent features. $B_{r}$ is ideal for this purpose as it contains the most recent data. 

\subsubsection{Reward Function}
The reward function $r^n$ is defined as,
\begin{equation} \label{eq:rn_pean}
  r^n=\begin{cases}
                e^n - |\hat{\mu} - \nu_1^n| \hspace{.5em} \text{if} \hspace{.5em} \nu_1^n < V_1\\ 
                e^n - |\hat{\mu} - \nu_1^n| \hspace{.5em} \text{if} \hspace{.5em} \nu_1^n > V_2\\ 
               e^n; \hspace{.5em} \text{Otherwise},
            \end{cases}
\end{equation}
\begin{equation} \label{eq:reward}
\text{where} \qquad \quad e^n =(1-(\text{L}_{c}^n-\text{L}_{c}^{n-1}))\times(1-\text{L}_{c}^n)
\end{equation}
and $V_1$ and $V_2$ are predefined thresholds.
$e^n$ is specified so that the reward will be higher for lower errors and higher rates of error change (Equation ~\ref{eq:reward}). Equation~\ref{eq:rn_pean} penalises $r^n$ if the network grows too large or too small.
\subsection{RA-DAE Algorithm}

With $S$, $A$ and $r^n$ defined, we present the general approach used to solve the MDP (Algorithm ~\ref{algo_qlearn}). Q-Learning was utilised with the following steps,\\

\begin{algorithm}[t]
\caption{RA-DAE algorithm}
\label{algo_deepmdpnet}
\begin{algorithmic}[1]
\Procedure{RA-DAE}{}
\State \textbf{define} : $n$ - Current batch ID
\State Initialise $Q(s,a)$ = 0 $\forall s \in S, a \in A $
\State $s,a=null$
\While{$\mathcal{D}_n \neq NULL$} 		
    \State $s',a',Q'$,$\Delta$Mrg,$\Delta$Inc,= GetCtrlParam($n,Q,s,a$)
    \If {$a'$ = Pool}
    	\State Fine-tune using $B_{ft}$
    \ElsIf {$a'$ = Increment}
    	\State Add $\Delta$Inc new nodes to the network
        \State Train the $\Delta$Inc nodes greedily using $B_{r}$
    \ElsIf {$a'$ = Merge}
    	\State Merge 2$\Delta$Mrg nodes into $\Delta$Mrg
    \EndIf
    \State Train the network with $\mathcal{D}_n$
    \State $s=s', a = a',Q=Q'$
    \State $n=n+1$
\EndWhile
\EndProcedure
\end{algorithmic}
\end{algorithm}

\noindent For the $n^{th}$ iteration, with data batch $\mathcal{D}^n$,
\begin{enumerate}
\item Until adequate samples are collected (i.e. $n \leq \eta_1$), train with B$_r$.
\item With adequate samples collected (i.e. $n>\eta_1$), start calculating Q-values for each state-action pair observed $\{s^n,a^n\}$, where $s^n \in S$, and $a^n \in A$ as defined in Algorithm ~\ref{algo_qlearn}.
\item During $\eta_1 < n \leq \eta_2$, uniformly perform actions from $A=\{$\emph{Increment},\emph{Merge},\emph{Pool}$\}$ to develop a fair utility estimate for all actions in $A$.
\item With an accurate estimation of $Q$ (i.e. $n>\eta_2$), the best action $a'$ is selected by $a'=argmax_{a'}(Q(s^n,a'))$ with a controlled amount of exploration ($\epsilon$-greedy).
\item if $a'=$\emph{Increment}, calculate $\Delta$Inc from Equation ~\ref{eq:inc}, add randomly initialised $\Delta$Inc nodes and greedily initialise \emph{only} the new nodes with B$_r$, while keeping the rest constant.
\item if $a'=$\emph{Merge} calculate $\Delta$Mrg from Equation ~\ref{eq:mrg} and average the closest pairs of $\Delta$Mrg nodes to amalgamate 2$\Delta$Mrg nodes to $\Delta$Mrg nodes.
\item if $a'=$\emph{Pool} fine-tune the network with B$_{ft}$.
\item Train the network with $\mathcal{D}^n$.
\item Calculate the new state, $s^{n+1}$ (Equation ~\ref{eq:cont-states}) and the reward $r^{n}$ (Equation ~\ref{eq:reward}).
\item Update the Value (Utility) Function $Q(s,a)$ as, 
\begin{equation} \label{eq:value_update}
Q^{(t+1)}(s^{n-1},a^{n-1}) = (1-\alpha)\times Q^{t}(s^{n-1},a^{n-1}) \\
+ \alpha \times q,
\end{equation}
where $q=r^n + \gamma \times max_{a'}(Q^{t}(s^n,a'))$.
\end{enumerate}

\begin{figure}[t]
\centering
	\includegraphics[width=0.45\textwidth]{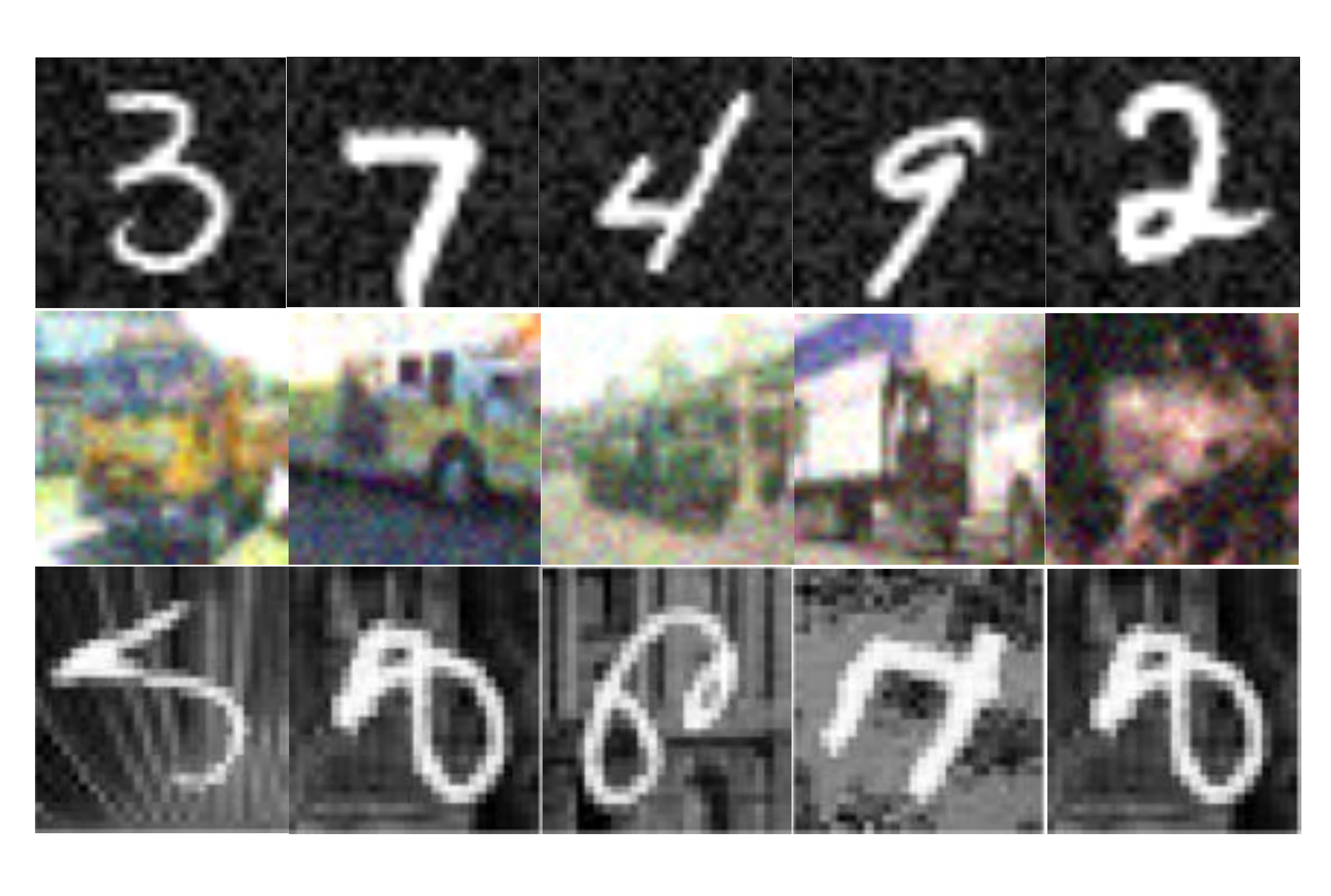}
    \vspace{-0.5cm}
	\caption{Random examples from the extended MNIST, CIFAR-10 and MNIST-rot-back datasets, respectively}
    \label{fig_mnist_noised}
\end{figure}

\vspace{-0.3cm}
\subsection{Function Approximation for Continuous Space}
\label{sub_sec_func_appx}
For clarity of presentation, Algorithms~\ref{algo_deepmdpnet} and ~\ref{algo_qlearn} assume discrete state space. However, the same algorithms can be extended for continuous state space. The idea is to, for a given action $a$ and an unseen state $\tilde{s}$, predict the utility value $Q(\tilde{s},a)=\hat{f}(\tilde{s},\mathbf{w})$ through function approximation where $\hat{f}$ is the function and $\mathbf{w}$ is the approximated parameter vector~\cite{sutton1998reinforcement}. In this paper, Gaussian Process Regression (GPR)~\cite{rasmussen2006gaussian} with squared exponential kernel, $k_{SE}(x,x')=\sigma^2(exp(-\frac{(x-x')^2}{2l^2}))$ has been used for this regression task. The hyperparameters $\sigma$ and $l$ are optimised by maximising the marginal likelihood w.r.t. the hyperparameters~\cite{rasmussen2006gaussian}. Formally, we collect at least $\eta_2 - \eta_1$ observed states and corresponding value pairs $\{s^n, Q(s^n,a^n)\}$. Next, for each $a \in A$, separate curves are fitted with GPR for the $\{s^n,Q(s^n,a^n)\}$ collection of pairs by separating pairs w.r.t. $a^n$, so that there are $|A|$ curves. Then, for an unseen state $\tilde{s}$ and a given action $a'$, $Q(\tilde{s},a')$ is calculated using the curve fitted for action $a'$. The continuous space is preferred as it provides a detailed representation of the environment with fewer variables, as opposed to the discrete space. This is sensible as the information extracted is continuous (e.g. $L_{g}, L_{c}, \nu$). 

\vspace{-0.3cm}
\subsection{Summary} 
Our proposed solution is detailed in Algorithm \ref{algo_deepmdpnet} and can be seen is a repeated application of Algorithm ~\ref{algo_qlearn}. For each batch of data $\mathcal{D}^n$, the state $s^{n+1}$ and reward $r^{n}$ is calculated using Equations ~\ref{eq:cont-states} and ~\ref{eq:reward} respectively. Next, the best action $a'$ for the new state is retrieved by $a'=argmax_a'(Q(s^{n+1},a')$. To calculate $Q(s^{n+1},\hat{a})$ for some action $\hat{a}\in A$, GPR is employed as explained in Section ~\ref{sub_sec_func_appx}. Next the action $a'$ is performed. Then the network is fine-tuned using $\mathcal{D}^n$. This process is repeated until the end of the data stream.

\begin{algorithm}[t]
\caption{Control Parameter Calculation algorithm}
\label{algo_qlearn}
\begin{algorithmic}[1]
\Procedure{GetCtrlParam($n,Q,s,a$)}{}
\State \textbf{define} : $n$ - Current batch ID
\State \textbf{define} : $Q$ - Utility function
\State \textbf{define} : $s,a$ - Previous state,action
\State \textbf{define} : $\gamma$ - Discount rate
\State \textbf{define} : $\alpha$ - Learning rate
\If {$n < \eta_1$}
    \State \Return $null,Pool,Q,0,0$
\EndIf
\State Calculate current state, $s'$ (Equation ~\ref{eq:cont-states})
\If {$s,a \neq null$}
	\State $q = r^n + \gamma \times max_{a'}(Q(s',a'))$
    \State $Q(s,a) = (1-\alpha) \times Q(s,a) + \alpha \times q$
\EndIf
\If {$n < \eta_2$}
	\State Evenly chose action $a'$ from $\in A$
\Else
	\State Explore with $\epsilon$-greedy ($\epsilon=0.1$)
    \State \textbf{OR}
    \State $a' = argmax_{a'}(Q(s',a'))$
\EndIf
\State Calculate $\Delta$Mrg and $\Delta$Inc (Eq.~\ref{eq:inc} and ~\ref{eq:mrg})
\State \Return $s',a',Q,$ $\Delta$Mrg,$\Delta$Inc
\EndProcedure
\end{algorithmic}
\end{algorithm}

%
\vspace{-0.2cm}
\section{Experiments} 

\subsection{Overview and Setup}
The experiments are based on extended versions of three datasets (i.e. MNIST\footnote{http://yann.lecun.com/exdb/mnist/}, MNIST-rot-back\footnote{www.iro.umontreal.ca/~lisa/twiki/bin/view.cgi/Public/MnistVariations} and CIFAR-10 \footnote{http://www.cs.toronto.edu/~kriz/cifar.html}). Random samples from each dataset are depicted in Figure ~\ref{fig_mnist_noised}. The extended versions of each dataset consist of 1,000,000 examples. Examples were masked with noise during the generation to make them unique. We generated non-stationary distributions for each dataset using Gaussian processes (GP)~\cite{rasmussen2006gaussian}, simulating the covariate shift effect. Formally, the ratio for each class of labels is generated using $ratio_k(t)=\frac{exp\{a_k(t)\}}{\sum_{j=1}^K exp\{a_j(t)\}}$ where $a_k(t)$ is a random curve generated by the GP.

Experiments were conducted with three different types of deep architectures; SDAE (Standard Denoising Autoencoders), MI-DAE (Merge-Incremental Denoising Autoencoders) and RA-DAE (our approach). For MI-DAE, we used a modified "update rule I" introduced in ~\cite{zhou2012supplementary} as they claim the performance is fairly robust to different update rules as follows, 
\[
  \Delta \text{N}_{t+1}=\begin{cases}
               \Delta \text{N}_t + 30; \hspace{.5em} \text{,} \hspace{.5em} \frac{\text{e}_t}{\text{e}_{t-1}} < (1-\epsilon _1)\\ 
               \Delta \text{N}_t/2; \hspace{.5em} \text{,} \hspace{.5em} \frac{\text{e}_t}{\text{e}_{t-1}} > (1-\epsilon _2) \\
               \Delta \text{N}_t, \hspace{.5em} \text{Otherwise}
            \end{cases}
\]
$\Delta$Mrg = $\ceil{\gamma_{ratio} \Delta \text{Inc}}$; for $\gamma_{ratio}=0.2$, as these modifications produced better performance.

Several initial layer configurations (hidden layer sizes) were used, as outlined in Table ~\ref{tbl_node_counts}. To refer to a certain algorithm, we use the following notation. We use the superscript for the number of layers and the subscript to indicate the size of each layer. For example, SDAE$^{l3}_{1500}$ denotes a SDAE with three layers and 1500 nodes in each layer. The configurations in Table ~\ref{tbl_node_counts} maximise the performance of the algorithms tested. The continuous state space (Equation ~\ref{eq:cont-states}) was used for all the experiments. We define two error measures for evaluating performance. A local error E$_{lcl}=L^{n+1}_{c}$, measured on a validation set, $\mathcal{D}^{n+1}$ (batch succeeding the current batch) and a global error E$_{glb}=\frac{\sum_{\forall i}L^i_{c}}{|\mathcal{D}_{test}|} \text{ s.t. } \mathcal{D}^i \in \mathcal{D}_{test}$ measured on an unseen independent test set $\mathcal{D}_{test}$, which contains an approximate uniform distribution of all the classes. These two sets of data enable us to respectively, evaluate how the network preserve immediate past knowledge and the globally accumulated knowledge.

\begin{figure}
\centering
	\includegraphics[width=0.5\textwidth]{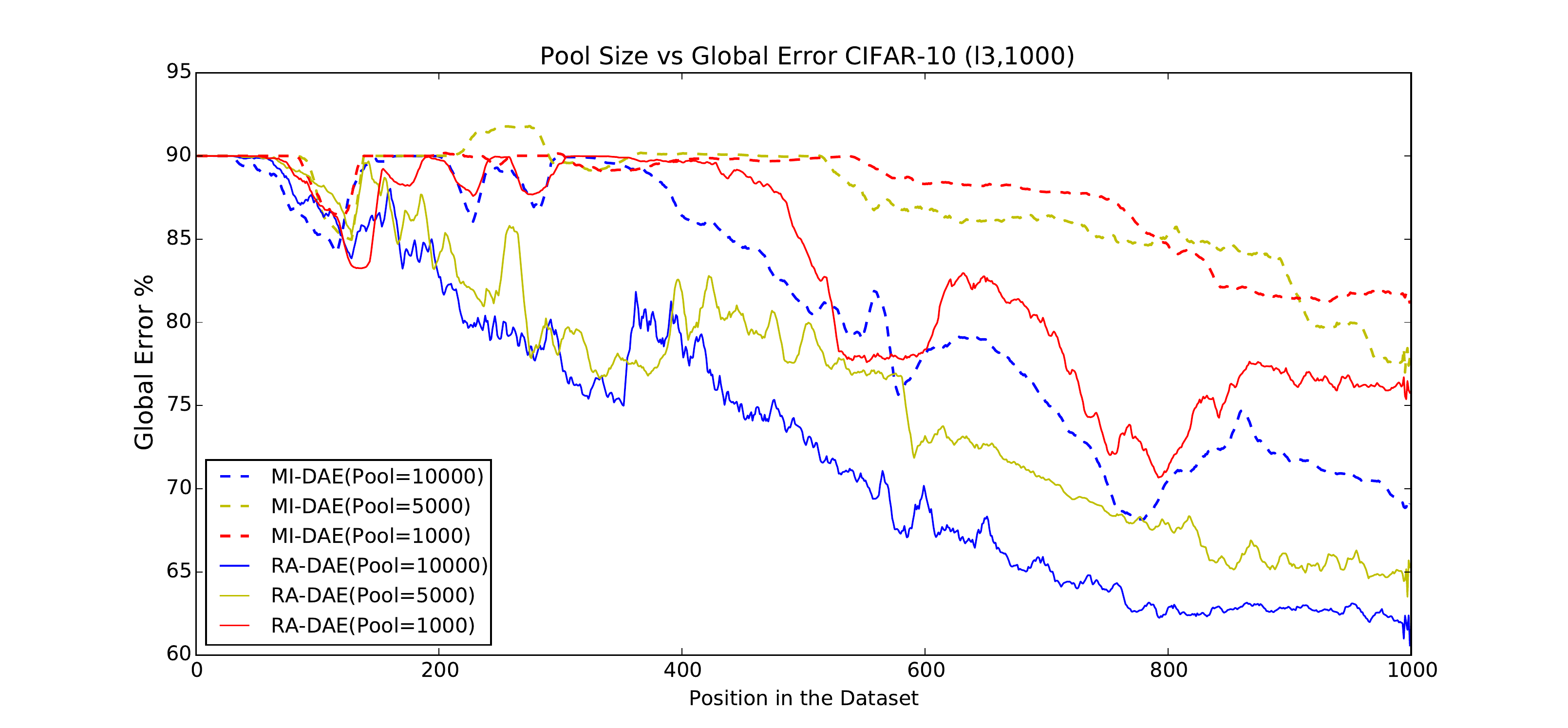}
    \vspace{-0.7cm}
	\caption{Performance of RA-DAE and MI-DAE for different pool sizes. Pool size of 10000 yielded the best results. It can be seen that RA-DAE with a pool size of 1000 performs similarly to MI-DAE with pool size 10000. This can be attributed to the learned policy and the pooling technique.}
    \label{fig_pool_size}
\end{figure}

All experiments were carried out using a Nvidia GeForce GTX TITAN GPU and Theano\footnote{http://deeplearning.net/software/theano/}. For all experiments we used 20\% corruption level, 0.2 learning rate, batch size of 1000. We empirically chose $\gamma=0.9$ (Equation ~\ref{eq:value_update}) $m$ (Equation ~\ref{eq:cont-states}) 30, and $\eta_1$ and $\eta_2$ (Algorithm ~\ref{algo_qlearn}) to be 30 and 60 respectively. $\Lambda$ (Equation ~\ref{eq:b_ft}) was selected as $0.7$ and $0.995$ for non-stationary and stationary experiments respectively. $\tau=10,000$ (for $B_r$, $B_{ft}$ and $B$) was chosen from a set of sizes \{1000, 5000, 10000\} as 10,000 produced the best results. Results are depicted in Figure ~\ref{fig_pool_size}.
\vspace{-0.3cm}

\subsection{Results}
\subsubsection{Evaluation of State Spaces}
\label{sub_sub_sec_eval_state}
\begin{figure}[t]
\centering
\includegraphics[width=0.46\textwidth]{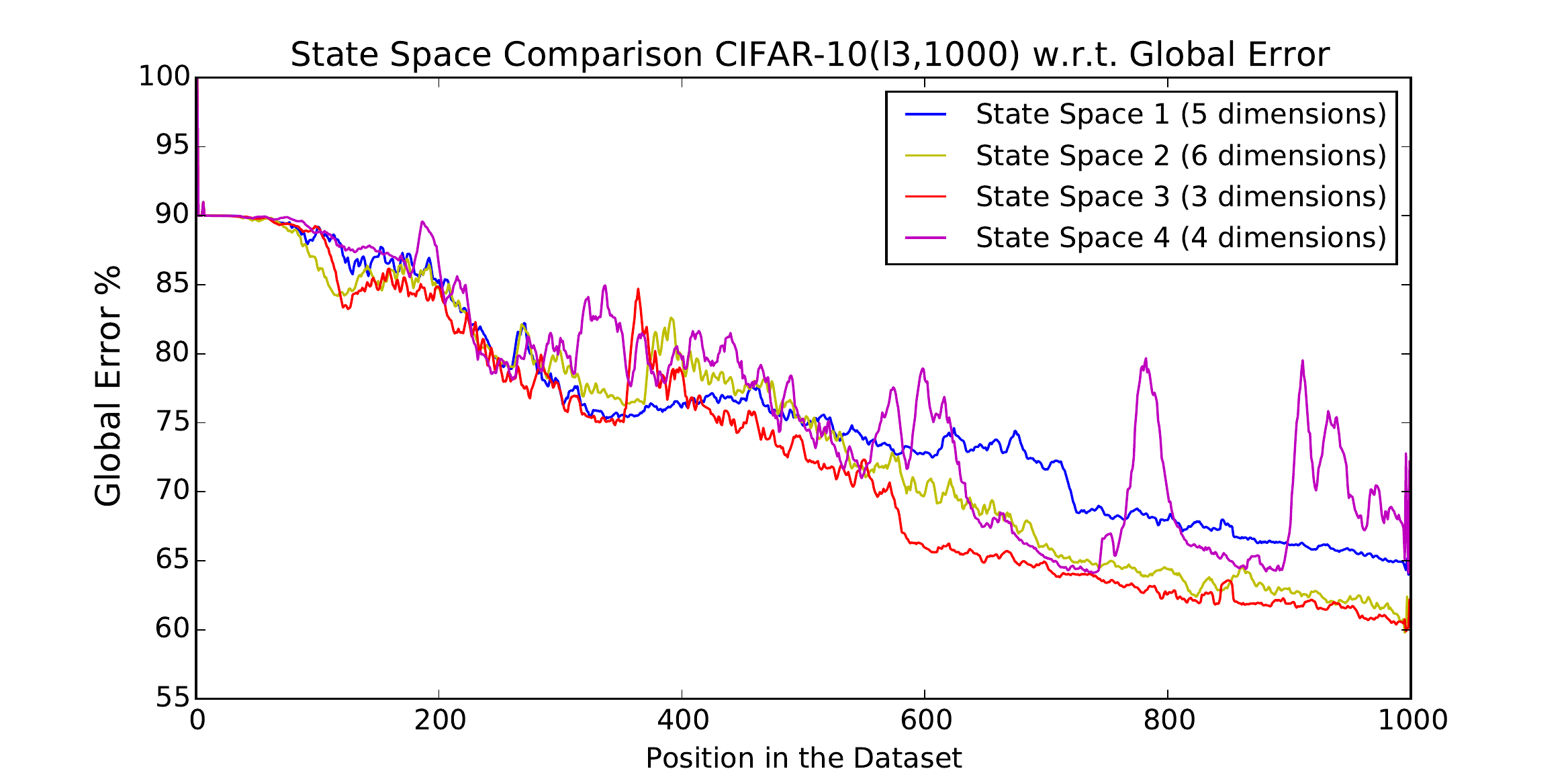}
\vspace{-0.3cm}
\caption{Analysis of Global Error E$_{glb}$ for different state spaces for CIFAR-10 and a network with 3 layers with 1000 neurons on each. The mathematical definitions of state spaces 1,2,3 and 4 can be found in Section \ref{sub_sub_sec_eval_state}. It is clear that State Space 4 shows a steeper reduction of error compared to its counterparts.}
\label{fig_state_comp}
\vspace{-0.3cm}
\end{figure}

\begin{figure*}
\centering
\hspace*{-1.0cm} 
\includegraphics[width=1.1\textwidth]{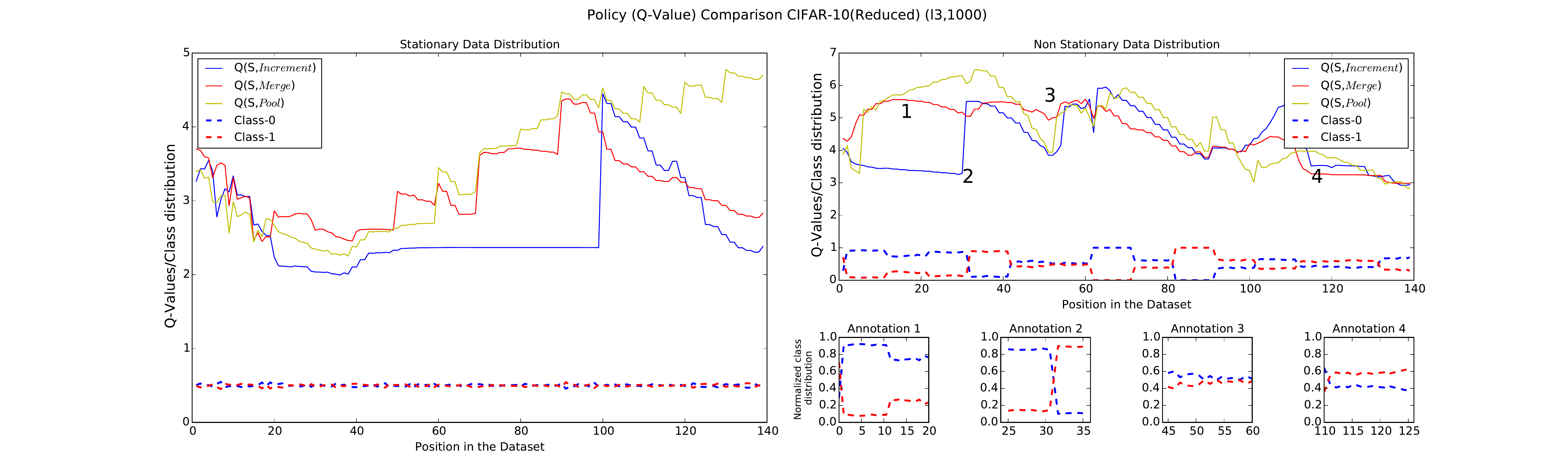}
\vspace*{-0.9cm}
\caption{Visualisation of the value function ($Q(s,a)$) evolution for stationary and non-stationary distributions. Annotations on the top-right graph indicate note-worthy behaviours of the value function. The left and top-right graphs depict the complete progression of the data distribution. For clarity, a reduced version of CIFAR-10 with only two classes and 200,000 examples was used. For stationary data distribution, the graph indicates how \emph{Pool} and \emph{Merge} operations dominate the behaviour as there are no significant data distribution changes. In the non-stationary setting, the value function for action \emph{Increment} surges in face of a sharp distribution change (Annotation 2). \emph{Merge} and \emph{Pool} operations take over when data distributions are consistent (Annotation 3 and 4 respectively).}
\label{fig_policy}
\end{figure*}

\begin{table}
\centering
\vspace{-0.2cm}
\caption{Initial layer configurations for different datasets. The superscript of the algorithm name specifies the number of layers and the subscript indicates the size of each layer. }
\begin{tabular}{ |c|c|c| }
 \hline
MNIST &CIFAR-10 & MNIST-rot-back \\
\hline
SDAE$^{l1}_{500}$ & SDAE$^{l1}_{1000}$ & SDAE$^{l1}_{1500}$ \\
SDAE$^{l3}_{500}$ & SDAE$^{l3}_{1000}$ & SDAE$^{l3}_{1500}$ \\
\hline
MI-DAE$^{l1}_{500}$ & MI-DAE$^{l1}_{1000}$ & MI-DAE$^{l1}_{1500}$\\
MI-DAE$^{l3}_{500}$ & MI-DAE$^{l3}_{1000}$ & MI-DAE$^{l3}_{1500}$\\
\hline
RA-DAE$^{l1}_{500}$ & RA-DAE$^{l1}_{1000}$ & RA-DAE$^{l1}_{1500}$\\
RA-DAE$^{l3}_{500}$ & RA-DAE$^{l3}_{1000}$ & RA-DAE$^{l3}_{1500}$\\
\hline
\end{tabular}
\label{tbl_node_counts}

\end{table}

As mentioned in Section ~\ref{sub_sub_sec_state_space} the state space was chosen while paying close attention to the performance against an unseen data batch, difference between observed data distributions and complexity of the network. We utilised various quantifiable measures. L$_{g}^{n+1}$ and L$_{c}^{n+1}$ were employed to evaluate RA-DAE's ability to classify an unseen batch of data (i.e. $\mathcal{D}^{n+1}$). Kullback-Leibler divergence ($D_{KL}(P^n||Q^n)$)~\cite{kullback1951information} was used to measure the divergence between the distribution of current data and previously fed data; $D_{KL}(P^n||Q^n)=\sum_i^K P^n(i)log(\frac{P^n(i)}{Q^n(i)})$, where $P^n(i)=\frac{Count_i^n}{p}$, $Count_i^n$ is the number of data points with class $i$ in $\mathcal{D}^n$, $p$ is as defined in Table ~\ref{tbl_def} and $Q^n(i)=\frac{\sum_{j-m}^j P^j(i)}{m}$. Finally the complexity of RA-DAE at a given time is captured by $\nu^n$.

With the aforementioned quantities defined, the following state spaces were defined:
\begin{itemize}
\item State Space 1 - $\{\mathcal{\tilde{L}}_g(m_3),\mathcal{\tilde{L}}_c(m_1),\mathcal{\tilde{L}}_c(m_2),\mathcal{\tilde{L}}_c(m_3),\nu\}$
\item State Space 2 - $\{\mathcal{\tilde{L}}_g(m_3),\mathcal{\tilde{L}}_c(m_1),\mathcal{\tilde{L}}_c(m_2),\mathcal{\tilde{L}}_c(m_3),\nu$,$D_{KL}(P^n||Q^n)\}$
\item State Space 3 - $\{\mathcal{\tilde{L}}_g(m),\mathcal{\tilde{L}}_c(m),\nu\}$
\item State Space 4 - $\{\mathcal{\tilde{L}}_g(m),\mathcal{\tilde{L}}_c(m),\nu$,$D_{KL}(P^n||Q^n)\}$
\end{itemize}

The constants $m_1,m_2,m_3$ and $m$ were chosen empirically and set to 5,15,30 and 30 respectively. The reason for calculating $\mathcal{\tilde{L}}$ for several $m$ values is to learn whether augmenting the state space of L$_g$ and L$_c$ contribute additional information. However, from the experimental results, it was evident that a simpler state space yields the best results. Furthermore, it was surprising to verify that $D_{KL}(P^n||Q^n)$ had no significant positive impact on the results. The performance of different state spaces is depicted in Figure ~\ref{fig_state_comp}.

\vspace{-0.1cm}
\subsubsection{Analysis of Structure Adaptation}
\begin{figure*}
\centering
\vspace{-0.2cm}
\includegraphics[width=1.0\textwidth]{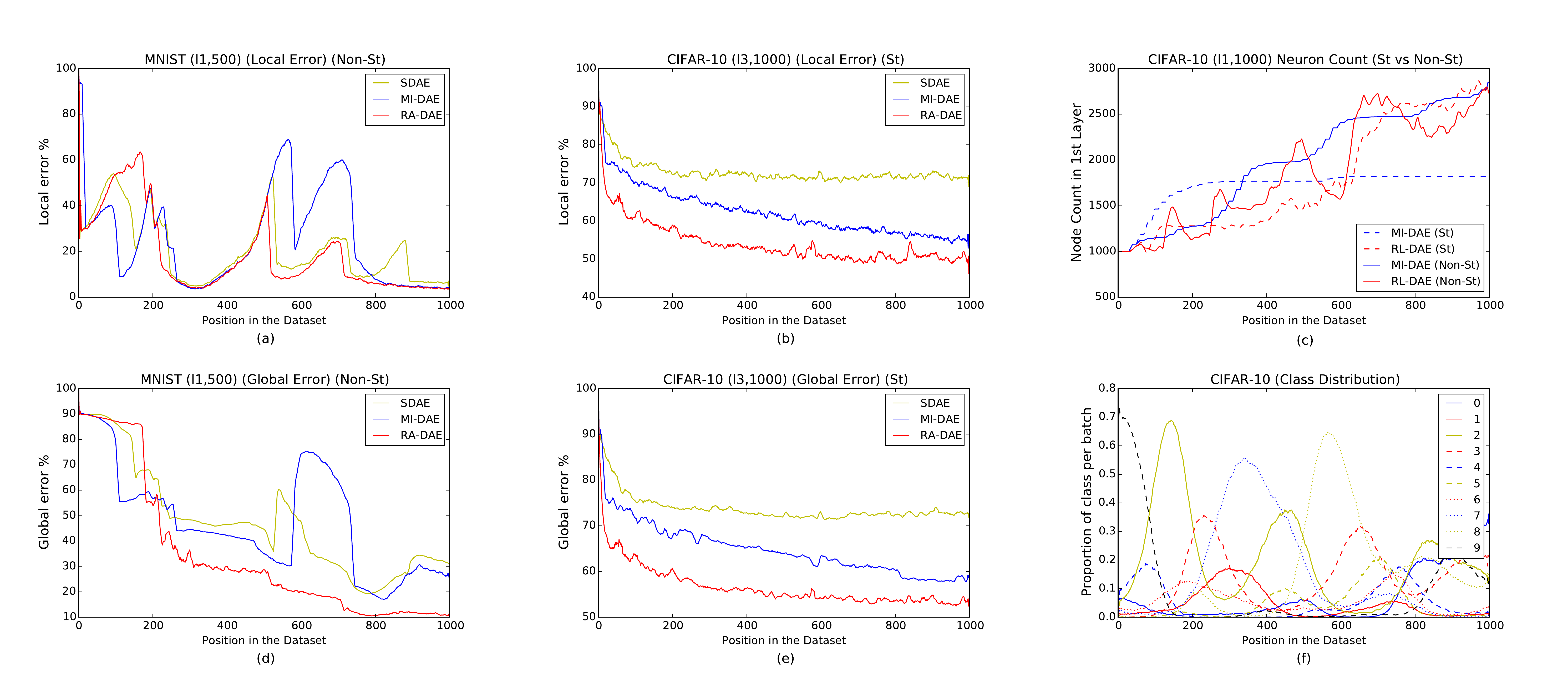}
\vspace{-0.7cm}
\caption{(a) and (d) show the behaviour of E$_{lcl}$ and E$_{glb}$ in a non-stationary (Non-St) situation, where (b) and (e) show the behaviour of E$_{lcl}$ and E$_{glb}$ in a stationary (St) situation. The titles consist of the name of the dataset followed by number of hidden layers and neuron count in each layer, within parenthesis. RA-DAE exhibits the lowest E$_{lcl}$ and E$_{glb}$ at the end, and a more consistent reduction compared to SDAE and MI-DAE. (c) presents node adaptation patterns of MI-DAE and RA-DAE in both stationary and non-stationary situations. (f) shows the class distribution of data over time and each curve denotes a single class. By comparing to (f), (c) clearly indicates that RA-DAE is more sensitive to changes in data distribution than MI-DAE in terms of the neuron adaptation. The horizontal axis represents the number of batches in the training dataset.}
\label{fig_test_plots}
\end{figure*}

We studied the adaptation pattern of RA-DAE and MI-DAE in both stationary and non-stationary environments. Figure \ref{fig_test_plots}(c) depicts the number of nodes in the first layer for MI-DAE and RA-DAE as they adapt to data distributions changes with the CIFAR-10 dataset. In non-stationary problems, RA-DAE exhibits repeated peaks in the number of nodes. This can be explained by the changes in class distribution in Figure ~\ref{fig_test_plots}(f). Node number changes in Figure \ref{fig_test_plots}(c) align with the peaks appearing for various class distributions in Figure ~\ref{fig_test_plots}(f). For sharp distribution changes, RA-DAE quickly increases the number of neurons. However, MI-DAE shows a moderate growth in the number of nodes, despite the rapid changes in the data distribution. This demonstrates that RA-DAE is more responsive than MI-DAE in adapting the architecture in the face of changes. For a stationary data distribution, MI-DAE shows a constant node count after the first few hundred batches, where RA-DAE increases the number of nodes over time. This can be attributed to the fact that reducing the number of neurons tends to increase the error,  occasionally making the reduction operation not preferable to RA-DAE. This is acceptable as RA-DAE will not increase nodes unnecessarily as it would lead to poor results due to \emph{overfitting}. An alternative is to perform the pool operation after reduce, which would reduce the error at an increased computational cost.

\begin{table*}[ht]
\centering
\caption{This table presents the E$_{lcl}$ and E$_{glb}$ obtained for various datasets and depths. Errors are in the format of mean$\pm$standard deviation for the last 250 batches. The lowest errors are highlighted in bold. RA-DAE has shown the best performance (smallest local and global errors) in most occasions (for both stationary and non-stationary). }
\vspace{-0.2cm}
\begin{tabular}{ |c|c|c|c|c|c|c||c|c|  }
 \hline
 &\multicolumn{2}{ |c| }{MNIST} &\multicolumn{2}{ |c| }{CIFAR-10} &\multicolumn{2}{ |c|| }{MNIST-rot-back} &\multicolumn{2}{ |c| }{CIFAR-10 (Stationary)}\\
  \hline
&E$_{lcl}$\% & E$_{glb}$\% &E$_{lcl}$\% & E$_{glb}$\% &E$_{lcl}$\% & E$_{glb}$\% &E$_{lcl}$\% & E$_{glb}$\%\\
 \hline
 SDAE$^{ l1}$   & $10.9 \pm 5.8$ & $27.2 \pm 5.7$ & $65.9\pm4.9$ & $82.8\pm1.1$ & $\mathbf{52.8\pm7.0}$ & $\mathbf{65.7\pm2.7}$ & $67.9\pm1.5$ & $70.2\pm0.6$ \\
 MI-DAE$^{ l1}$&   $6.4\pm 3.2$  & $23.9 \pm 4.4$  & $\mathbf{50.4\pm4.7}$ & $\mathbf{74.9\pm3.0}$ & $61.8\pm9.0$& $72.0\pm2.6$& $55.5\pm1.9$ & $\mathbf{61.4\pm0.8}$ \\
 RA-DAE$^{ l1}$& $\mathbf{5.1\pm1.4}$ & $\mathbf{11.3\pm0.7}$  & $\mathbf{50.6\pm7.2}$ & $\mathbf{74.0\pm2.4}$  &  $62.3\pm8.8$ & $69.6\pm2.8$ & $59.8\pm1.9$ & $\mathbf{61.9\pm1.1}$\\
 \hline
 SDAE$^{ l3}$ &$11.2 \pm 5.9$ & $31.6 \pm 5.4$ & $76.3\pm6.6$ & $88.4\pm1.9$ & $67.6\pm8.9$& $77.1\pm3.2$ & $71.8\pm1.4$ & $72.7\pm0.7$\\
 MI-DAE$^{ l3}$& $5.4 \pm 4.4$ & $31.3\pm4.0$ & $43.7\pm8.5$ & $71.0\pm1.6$  & $56.0\pm9.2$ & $65.5\pm2.1$ & $56.1\pm1.8$ & $58.9\pm1.1$\\
 RA-DAE$^{ l3}$& $\mathbf{4.1 \pm 3.0}$ & $\mathbf{13.4\pm0.1}$ & $\mathbf{32.4\pm8.0}$ & $\mathbf{62.7\pm0.7}$ & $\mathbf{48.2\pm9.2}$ & $\mathbf{60.6\pm3.4}$ & $\mathbf{50.6\pm2.1}$ &$\mathbf{53.6\pm2.1}$ \\
 \hline
\end{tabular}
\label{tbl_errors}
\end{table*}

\vspace{-0.1cm}
\subsubsection{Analysis of Local and Global Error}
Finally, the capability to preserve past knowledge, balancing immediate and global rewards for the algorithms was assessed by using the local error, E$_{lcl}$, and the global error, E$_{glb}$. We used the hybrid objective function ($L_{disc}+\lambda L_{gen}$ for $\lambda=0.2$) ~\cite{zhou2012online} to fine-tune the network.

Figure ~\ref{fig_test_plots} depicts several interesting results. Figure ~\ref{fig_test_plots}(a) illustrates the behaviour of the E$_{lcl}$. RA-DAE$^{l1}_{500}$ shows a clear improvement w.r.t E$_{lcl}$ over time. Note how in RA-DAE$^{l1}_{500}$ the fluctuations shrink over time. Moreover, Figure~\ref{fig_test_plots}(d) delineates a significant E$_{glb}$ error margin maintained by RA-DAE$^{l1}_{500}$ compared to SDAE$^{l1}_{500}$ and MI-DAE$^{l1}_{500}$. RA-DAE's ability to grow the network faster compared to MI-DAE explains this significantly lower error. Figure ~\ref{fig_test_plots}(b) and (e) portray the performance of the algorithm in a stationary environment (CIFAR-10). Though we expected all algorithms to perform comparably well in the stationary environment, RA-DAE$^{l3}_{1000}$ achieves the lowest E$_{lcl}$ and E$_{glb}$ and the steepest error reduction. Both RA-DAE$^{l3}_{1000}$ and MI-DAE$^{l3}_{1000}$ demonstrate better performance than SDAE$^{l3}_{1000}$. This highlights that structure adaptation strategies enhance the performance of deep networks in both stationary and non-stationary environments. 

Table~\ref{tbl_errors} summarises the errors (mean$\pm$standard deviation of the last 250 batches) for various datasets. The number 250 was chosen, as the last 250 batches displayed a consistent performance in most instances. There are several key observations from Table~\ref{tbl_errors}. First, RA-DAE$^{l3}$ has outperformed its counterparts in both stationary and non-stationary scenarios, where RA-DAE$^{l1}$ and MI-DAE$^{l1}$ have performed equally well. By observing the performance of RA-DAE$^{l1}$ and RA-DAE$^{l3}$ it is evident that the performance of RA-DAE has improved as the network becomes deeper. MI-DAE has exhibited the same property in most occasions. The rationale being, not only deep networks are more robust to structural modifications in terms of error, but also they are able to learn more descriptive representations as depth increases. However, performance of SDAE$^{l3}$ is worse than SDAE$^{l1}$ in both cases. This observation justifies the need for better techniques to leverage deep architectures in online scenarios. 

A surprising observation can be made in \{SDAE,MI-DAE,RA-DAE\}$^{l1}$ for MNIST-rot-back. Even though we expected RA-DAE to perform the best, SDAE$^{l1}$ shows the best performance with a $52.8\pm7.0$\% and $65.7\pm2.7$\% for E$_{lcl}$ and E$_{glb}$ respectively. Close examination of the behaviours of E$_{lcl}$ and E$_{glb}$ of SDAE, MI-DAE and RA-DAE, shows that MI-DAE and RA-DAE do not perform as well as SDAE. This is due to the fluctuation of E$_{lcl}$ being fast, which causes the algorithm to increase the number of nodes unnecessarily. Consequently, MI-DAE and RA-DAE lead to poor accuracy due to \emph{overfitting}. This issue alleviates as the network becomes deeper.

\vspace{-0.1cm}
\subsubsection{Analysis of the Policy Learnt}
In order to analyse the policy learnt by RA-DAE, it is imperative to take a close look at the value function (i.e. $Q(s,a)$) learnt by RA-DAE. Figure ~\ref{fig_policy} depicts the evolution of the value function over time with note-worthy behaviours annotated. For the purpose of visualisation, a simplified version of CIFAR-10 dataset (CIFAR-10-bin) has been used.  CIFAR-10-bin comprises only two classes and has a total of 200,000 data points. Figure \ref{fig_policy} depicts the value function for two settings; stationary and non stationary. The annotation graphs at top-right highlight the changes in data distribution at the points of interest in the top graph. 

In the stationary setting, it can be seen that \emph{Pool} and \emph{Merge} operations have dominated the policy, Figure ~\ref{fig_policy}(right). This is sensible as the data distribution stays constant throughout and a necessity to increase the number of nodes hardly emerges. 

For the non-stationary setting, it can be seen how \emph{Pool} and \emph{Merge} operations have a high value as the algorithm has not seen an significant data distributions, thus suppressing \emph{Increment} operation. Next, at annotation 2 it can be seen how the value of \emph{Increment} operation boosts up due to the massive data distribution change. Then, at point 3, \emph{Merge} operation takes over as data distribution is somewhat consistent. And finally, at point 4, \emph{Pool} operation dominates the graph due to the consistency of the distribution of data.


 

\section{Conclusion}

Online learning can be widely beneficial for deep architectures as it allows network adaptation for streaming data problems. However, defining the structure of the network, including number of nodes, can be difficult to do in advance. To address this, \cite{zhou2012online} introduces MI-DAE which can dynamically change the structure of the network but relies on simple heuristics. The novelty of this work is an online learning stacked denoising autoencoder which leverages reinforcement learning to modify the structure of the deep network. In this, we use a model-free reinforcement learning approach and calculate a utility function for actions by sampling from the incoming states. 

Compared to the counterpart, our approach is more principled and responsive in adapting to new information. The method leverages RL to make decisions in a dynamic fashion. The control behaviour combined with powerful pooling techniques allows our approach to preserve past-knowledge effectively. Finally, our solution make decisions based on long-term versus immediate reward. Experimental results indicate that our solution often outperforms its counterparts with a lower classification error, and the performance improves as the network becomes deeper. Also, the approach is more sensitive to changes in the data distribution. Future work will address other deep learning architectures such as convolutional neural nets and deep Boltzmann machines.

\section*{Acknowledgements}
This research was supported by funding from the Faculty of Engineering \& Information Technologies, The University of Sydney, under the Faculty Research Cluster Program. We gratefully acknowledge the support of NVIDIA Corporation with the donation of the GPU used for this research.

\newpage
\bibliography{ecai}
\end{document}